\pdfoutput=1

\documentclass[11pt]{article}

\usepackage{booktabs}
\usepackage{array}
\usepackage{times}
\usepackage{latexsym}
\usepackage{amsmath}
\usepackage{natbib}
\usepackage{bm}
\usepackage{relsize}
\usepackage{amssymb} 

\usepackage{microtype}

\usepackage{inconsolata}

\usepackage{CJKutf8}

\usepackage{graphicx}
\usepackage{subfigure}
\usepackage{multirow}

\usepackage{relsize}
\usepackage{graphicx}
\usepackage{makecell}
\usepackage[final]{acl}

\usepackage{times}
\usepackage{latexsym}

\usepackage[T1]{fontenc}

\usepackage[utf8]{inputenc}

\usepackage{microtype}

\usepackage{inconsolata}

\title{CMDAG: A Chinese Metaphor Dataset with Annotated Grounds as \\CoT for Boosting Metaphor Generation}

\author{
    \thanks{Equal contribution}Yujie Shao\textsuperscript{2}\quad
    \footnotemark[1]Xinrong Yao\textsuperscript{3}\quad
    \footnotemark[1]Xingwei Qu\textsuperscript{1,6}\\
    \textbf{Chenghua Lin}\textsuperscript{6}\quad
    \textbf{Shi Wang}\textsuperscript{8}\quad
    \textbf{Stephen W. Huang}\textsuperscript{9}\quad
    \thanks{Corresponding author.}\textbf{Ge Zhang}\textsuperscript{1,4,5,7}\quad
    \footnotemark[2]\textbf{Jie Fu}\textsuperscript{1,5}\\
    \textsuperscript{1}HKUST \quad \textsuperscript{2} University of California, San Diego \quad
    \textsuperscript{3}Massachusetts Institute of Technology\\
    \textsuperscript{4}University of Waterloo \quad
    \textsuperscript{5}Multimodal Art Projection Research Community \quad
    \textsuperscript{6}University of Manchester \\
    \textsuperscript{7}Stardust.ai \quad
    \textsuperscript{8}Institute of Computing Technology, Chinese Academy of Sciences \quad
    \textsuperscript{9}Harmony.ai\\
    \vspace{-4ex}
\small
\texttt{ge.zhang@uwaterloo.ca,jiefu@ust.hk} \\
}

\begin{document}
\maketitle
\begin{abstract}
Metaphor is a prominent linguistic device in human language and literature, as they add color, imagery, and emphasis to enhance effective communication. This paper introduces a large-scale high quality annotated Chinese Metaphor Corpus, which comprises around 28K sentences drawn from a diverse range of Chinese literary sources, such as poems, prose, song lyrics, etc. To ensure the accuracy and consistency of our annotations, we introduce a comprehensive set of guidelines. These guidelines address the facets of metaphor annotation, including identifying tenors, vehicles, and grounds to handling the complexities of similes, personifications, juxtapositions, and hyperboles. Breaking tradition, our approach to metaphor generation emphasizes grounds and their distinct features rather than the conventional combination of tenors and vehicles. By integrating "ground" as a CoT (Chain of Thoughts) input, we are able to generate metaphors that resonate more with real-world intuition.
We test generative models such as Belle, Baichuan, and Chinese-alpaca-33B using our annotated corpus. 
These models are able to generate creative and fluent metaphor sentences more frequently induced by selected samples from our dataset, demonstrating the value of our corpus for Chinese metaphor research. 
The code is available in \url{https://github.com/JasonShao55/Chinese_Metaphor_Explanation}.
\end{abstract}

\begin{CJK*}{UTF8}{gbsn}
\section{Introduction}

\begin{figure}
    \centering
    \includegraphics[width=0.8\linewidth]{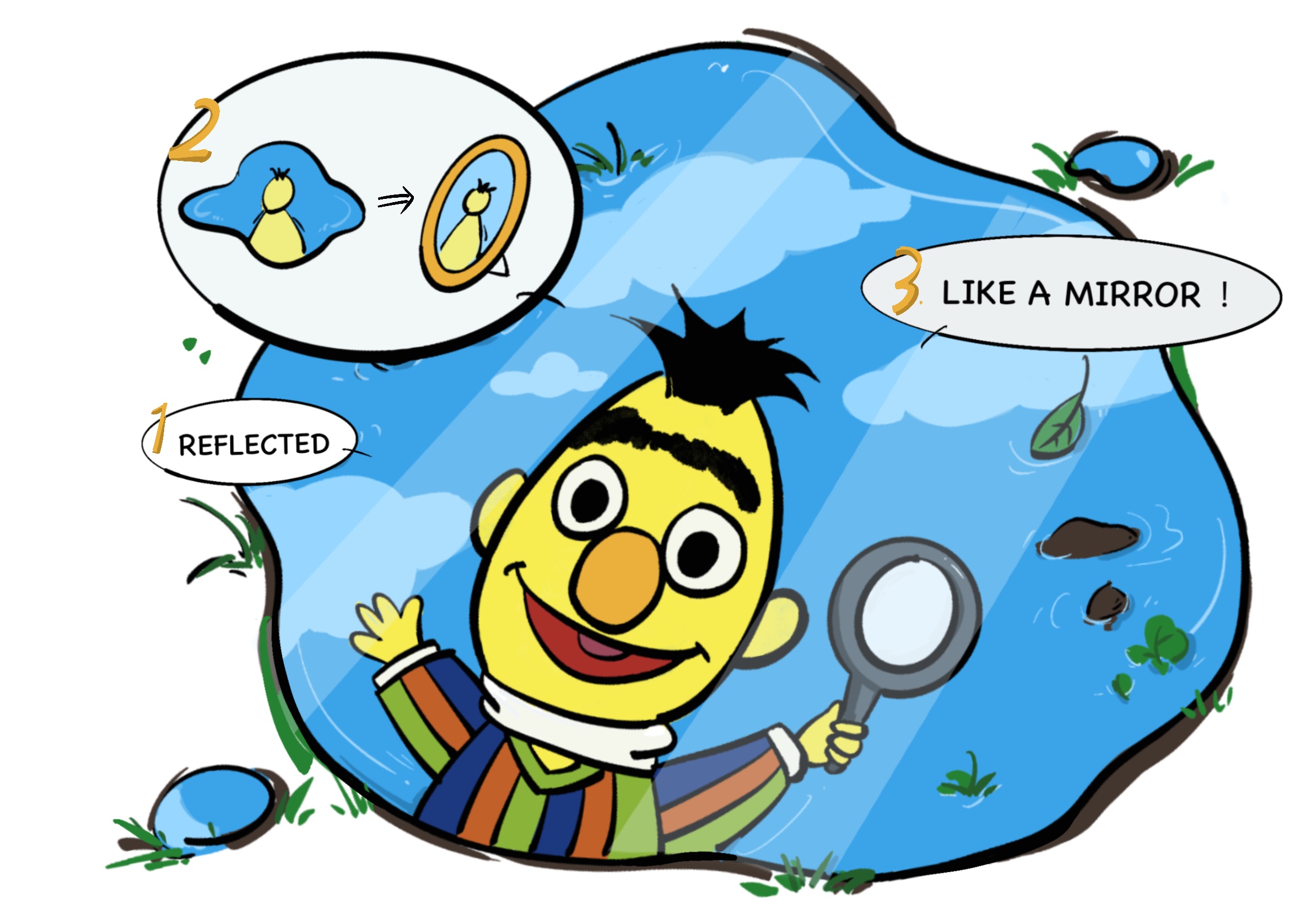}
    \caption{Sketch Map of the Metaphorical Language Writing Process.}
    \label{fig:sketch_map}
\end{figure}

Metaphor is a prominent linguistic device in human language and literature, typically to draw a comparison between disparate objects or concepts with the intent to make the expression more vivid, or make abstract concepts easier to understand.

With the progression of computational linguistics, there is an increasing focus on metaphor generation through machine learning techniques, notably in chatbot applications.  \citet{zheng2020love} shows how machine-generated nominal metaphors (NMs) can significantly enhance user interest during interactions with chatbots.
\citet{li2022nominal} finds substantial applications in shaping downstream task outputs in Natural Language Generation (NLG). Notably, a series of evaluations by  \citet{chakrabarty2020generating,chakrabarty2021mermaid} conduct human evaluations comparing literal expressions from machine-generated stories and poems with machine-generated metaphors and find users prefer the text with metaphors.

However, metaphors are referred to as novel linguistic expressions where an object or concept is used outside of its normal conventional meaning to express another meaning under a given context. Intrinsically, metaphors do not reside within the language itself but in the way they conceptually map one mental domain onto another in application \cite{lakoff1992}, as shown in Fig.~\ref{fig:sketch_map}. 

With this consideration, we establish metaphor sentences centered on identifying the conceptual mappings within metaphors, specifically \textsmaller{GROUNDS} (喻意). Metaphors consist of two components: \textsmaller{TENORS} (本体), representing the actual subject, and \textsmaller{VEHICLES} (喻体), symbolizing the comparative element. Employing \textsmaller{GROUNDS} can enhance sentence fluency and creativity, achieving human-like metaphorical expression ~\citep{yang-etal-2023-fantastic}. 
This work introduces an annotated Chinese metaphor corpus (\textbf{CMDAG}) that is derived from a diverse range of Chinese literature. Every metaphorical sentence within the corpus is accompanied by its corresponding \textsmaller{GROUNDS}. The central aim of our annotation effort is to accurately annotate each metaphor with a well-defined tuple of features (\textsmaller{TENOR}, \textsmaller{VEHICLE} and \textsmaller{GROUND}), capturing its intrinsic elements.

\begin{table*}
\centering
\setlength\tabcolsep{3pt}
\caption{Statistic characteristics and annotation information of main existing Chinese metaphor/simile datasets of metaphor and simile and CMDAG dataset. W and F separately denote the tenor/vehicle words and
the corresponding feature words.}
\label{tab:datasets}
\begin{tabular}{lccccccc}
\toprule
\textbf{Dataset} & \textbf{\# Nums} & \textbf{Tenor} & \textbf{Vehicle} & \textbf{Ground} & \textbf{Context} & \textbf{Open-source} \\
\cmidrule(lr){3-3} \cmidrule(lr){4-4} \cmidrule(lr){6-6}
 & & \textbf{W/F} & \textbf{W/F} &  & \textbf{Above/Below} &  \\
\midrule
Poetry~\citep{liu-etal-2019-rhetorically} & 43,051 & $-/-$  & $-/-$ & $-$ & $\checkmark/-$ & $\checkmark$\\
Lyrics~\citep{liu-etal-2019-rhetorically} & 246,669 & $-/-$  & $-/-$ & $-$ & $\checkmark/-$ & $\checkmark$\\
CS~\cite{zhang2021writing} & 5,490,721 & $-/-$  & $-/-$ & $-$ & $\checkmark/\checkmark$ & $\checkmark$\\
CMC~\cite{li-etal-2022-nominal} & 2,787 & $\checkmark/-$  & $\checkmark/-$ & $-$ & $-/-$ & $\checkmark$\\
GraCe~\cite{yang-etal-2023-fantastic} & 61,360 & $\checkmark/\checkmark$  & $\checkmark/\checkmark$  & $\checkmark$ & $\checkmark/\checkmark$ & $-$ \\
\midrule
CMDAG & 27,989 & $\checkmark/\checkmark$  & $\checkmark/\checkmark$& $\checkmark$ & $\checkmark/\checkmark$ & $\checkmark$\\
\bottomrule
\label{tab: Chinese Metaphor Corpora}
\end{tabular}
\end{table*}

To evaluate the effectiveness of our annotated corpus for Chinese metaphor generation, we undertake two evaluative setups,  both incorporating \textsmaller{GROUNDS} and the Chain of Thought (CoT) capability of language models. In the first setup, we prompt with \textsmaller{TENOR} and \textsmaller{VEHICLE}, and we allow the language model to deduce \textsmaller{GROUNDS}. For the second setup, we prompt \textsmaller{TENORS} and \textsmaller{GROUNDS}, and fine-tune metaphorical sentence generation by asking the language model to deduce a plausible \textsmaller{VEHICLE}. In summary, our paper outlines the following three contributions: 

\begin{enumerate}
    \item We present \textbf{CMDAG}, a unique Chinese metaphor dataset, wherein a key feature is the inclusion of \textsmaller{GROUNDS}. This dataset's thoughtful design enables the intuitive generation of metaphorical constructs, addressing a notable absence in contemporary research literature.

    \item We introduce a metaphor annotation pipeline by leveraging academically specialized annotators' expertise, achieving enhanced precision in pinpointing the \textsmaller{GROUNDS} of metaphors.
    
    \item We propose the first work introducing CoT into metaphor generations. Given \textsmaller{TENOR} and \textsmaller{VEHICLE}, deriving \textsmaller{GROUNDS} using CoT, language models can generate coherent and integrative metaphor sentences. Furthermore, by combining \textsmaller{TENORS} and \textsmaller{GROUNDS}, we enhance the generation of \textsmaller{VEHICLE}, improving the quality of the generated metaphorical expressions.

\end{enumerate}

\section{Related Work}
\subsection{Chinese Metaphor Corpora}
Metaphor is not only a literature of rhetoric but also a way of thinking rooted in Chinese culture~\cite{lin2021metaphor}.
However, due to the shortage of Chinese corpora~\cite{zhang2023chinese}, researchers are still in lack of high-quality Chinese metaphor corpora. 
Lyrics and Poetry corpora released by~\citep{liu-etal-2019-rhetorically} provide a great source of metaphorical Chinese language, but they do not dig in and provide fine-grained annotations of existing Chinese similes and metaphors in Lyrics and Poetry corpora.
CS~\cite{zhang2021writing}, another large Chinese rhetorical corpus, is in shortage of Fine-grained annotation as well.
CMC~\cite{li-etal-2022-nominal} is a valuable Chinese metaphor corpus with cautious annotation of tenors and vehicles, but CMC is pretty small and without the annotation of \textsmaller{GROUNDS} (喻意).
GraCe~\cite{yang-etal-2023-fantastic}, an amazing contemporaneous research work, claims to provide a carefully annotated Chinese simile corpus but hasn't been released yet, and only focuses on clearly stated Chinese similes.
As a sharp contrast, CMDAG is a carefully annotated large Chinese metaphor (also with simile) corpus with annotations of all \textsmaller{TENOR}, \textsmaller{VEHICLES}, and \textsmaller{GROUNDS}, which is a valuable resource for researchers interested in Chinese metaphor processing.
We briefly compare the existing major Chinese metaphor/simile corpora in Tab.~\ref{tab: Chinese Metaphor Corpora}.

\subsection{Boosting NLG via Chain-of-Thought}
Chain-of-Thought (\textbf{CoT}) is the most important inference trick inducing Large-scale Language Models (\textbf{LLMs}) to output reasonable results~\cite{wang2023interactive} since proposed by \citet{wei2022chain}.
It has been widely used in different LLM-based Natural Language Generation (\textbf{NLG}) tasks, including human moral value alignment~\cite{liu2023training, liu2022second}, math problem solving~\cite{yue2023mammoth}, and evaluation of NLG results~\cite{jiang2023tigerscore, chan2023chateval}.

As illustrated in Fig.~\ref{fig:sketch_map}, we believe that \textsmaller{GROUNDS} is the natural CoT connecting \textsmaller{TENOR} with \textsmaller{VEHICLE}, which has been discussed in literature research works~\cite{black1979more,end1986grounds} and NLP research works~\cite{gong2003corpus,stowe-etal-2021-metaphor,wachowiak2023does}.
Specifically, \citet{li2023metaphor} propose to introduce explicit basic meaning modeling to boost metaphor detection.
Additionally, \citet{yang-etal-2023-fantastic} reveal that simile generation could benefit from pre-specified constraints, especially explicitly stated \textsmaller{GROUNDS}.
As a sharp comparison, CMDAG directly verifies how LLMs perform on Chinese metaphor generation in various settings, especially with the assistance of \textsmaller{GROUNDS} (喻意) as CoT. 

\begin{table*}
\centering
\footnotesize 
\setlength\tabcolsep{5pt}
\begin{tabular}{lccc}
    \toprule
    \textbf{Source Type} & \textbf{\# Literature Works} & \textbf{\# Likely-Metaphors} & \textbf{\# Annotated Metaphors} \\
    \midrule
    Prose/Poem & 3,459 & 28,553 & 5,294 \\
    Song Lyrics & 102,197 & 109,827 & 21,276 \\
    Contemporary Poem & 4,494 & 7,268 & 939 \\
    HipHop/Rap Lyrics & 3,004 & 7,603 & 480 \\
    \midrule
    \textbf{Total} & 113,154 & 153,251 & 27,989 \\
    \bottomrule
\end{tabular}
\caption{\label{corpus_sources}
Statistics of CMDAG and its raw data collection literature sources.
}
\end{table*}

\section{Chinese Metaphor Dataset}

In this section, we present our annotated dataset of Chinese metaphors. Subsequent subsections establish basic definitions used in our dataset, and provide detailed insights into the data collection and annotation processes.

\subsection{Definition}
A \textbf{Metaphor (暗喻/隐喻)} is a linguistic device in which a word or phrase literally denoting one kind of object or idea is used in place of another to suggest a likeness or analogy between them.
For example, the metaphor "何等动人的一页又一页篇章！这是人类思维的花朵。" compares the tenor, the pages of literature (一页页篇章), to the vehicle, bloom of human thoughts (人类思维的花朵), to convey the beautiful nature of the literature's expressions.
In CMDAG, we uniformly formalize and process Chinese similes and metaphors for convenience, since similes are also sometimes referred to as direct metaphors\footnote{\href{http://www.vismet.org/metcor/documentation/relation_to_metaphor.html}{Relation to metaphor} from BNC Baby specifications.}.

To further explain other annotated elements, \textbf{Tenor (本体)} is the literal object or idea being described, and \textbf{Vehicle (喻体)} is the object or idea carrying the weight of comparison. The \textbf{Ground (喻意)} of a metaphor/simile is the concept or concepts the tenor and vehicle share, enabling the metaphor to align with common sense.
\begin{table*}
\centering
\scriptsize
\relsize{-0.5}
\setlength\tabcolsep{3pt}
\begin{tabular}{llccc}
    \toprule
    \textbf{Source Type} & \textbf{Sentence} & \textbf{Tenor} & \textbf{Vehicle} & \textbf{Ground}\\
    \midrule
    \multirow{9}{*}{Prose/Poem} 
    & 雨，像银灰色黏湿的蛛丝，织成一片轻柔的网，& \multirow{2}{*}{雨} & \multirow{2}{*}{蛛丝} & \multirow{2}{*}{细长的形状} \\
    & 网住了整个秋的世界。 \\
    & The rain is like silver-gray sticky spider silk, weaving & \multirow{2}{*}{rain} & \multirow{2}{*}{spider silk} & \multirow{2}{*}{elongated shape} \\
    & into a soft net that captures the entire realm of autumn. \\
    \cmidrule(l){2-5}
    & 佛法就好像手中的一块玉，如果没有握过许多泛泛的 & \multirow{2}{*}{佛法} & \multirow{2}{*}{玉} & \multirow{2}{*}{珍贵的属性} \\
    & 石头，就不能了解手中的玉是多么珍贵了。 \\
    & Buddhism is like a piece of jade in your hand, if you & \multirow{3}{*}{Buddhism} & \multirow{3}{*}{jade} & \multirow{3}{*}{preciousness} \\
    & have not held many ordinary stones, you cannot \\
    & understand how precious the jade is. \\
    \midrule
    \multirow{6}{*}{Song Lyrics}
    & 我以为旅人将我热情都燃尽\;你却像一张情书感觉很初级 & 你 & 情书 & 稚嫩的感情 \\
    & I thought travelers would burn out all my passion, & \multirow{2}{*}{you} & \multirow{2}{*}{love letter} & \multirow{2}{*}{immature emotion} \\
    & but you are like a love letter, which feels so elementary \\
    \cmidrule(l){2-5}
    & 爱像一阵风\;吹完它就走 & 爱 & 风 & 短暂的经过 \\
    & Love is like a gust of wind, it blows away and then & \multirow{2}{*}{love} & \multirow{2}{*}{wind} & \multirow{2}{*}{a brief passage} \\
    & goes away \\
    \midrule
    \multirow{5}{*}{Contemporary Poem}
    & 花香仿佛消散的钟声 & 花香 & 消散的钟声 & 浅淡的感觉 \\
    & The fragrance of flowers is like the dissipation of bells & fragrance of flowers & dissipated bells & light feeling \\
    \cmidrule(l){2-5}
    & 我的颊像溶了的雪，我的心像热了的酒 & 我的脸颊;我的心 & 溶了的雪; 热了的酒 & 温暖的感觉;炙热的感觉 \\
    & My cheeks are like melted snow, and my heart is & \multirow{2}{*}{my cheeks;my heart} & \multirow{2}{*}{melted snow;warm wine} & \multirow{2}{*}{warm feeling;hot feeling} \\
    & like warm wine \\
    \midrule
    \multirow{6}{*}{HipHop/Rap Lyrics}
    & 我的努力依旧还不够\;如同大海里的虾米 & 我 & 大海里的虾米 & 渺小的状态 \\
    & My efforts are still not enough, like shrimps in the sea & Myself & shrimps in the sea & insignificance \\
    \cmidrule(l){2-5}
    & 纵然现实残酷\;惟有向着上天祷告 & \multirow{2}{*}{生活} & \multirow{2}{*}{战场} & \multirow{2}{*}{激烈的斗争} \\
    & 生存就仿似生活响一个战场 \\
    & Even though the reality is cruel, we can only pray & \multirow{2}{*}{life} & \multirow{2}{*}{battlefield} & \multirow{2}{*}{fierce struggle } \\
    & to God. Survival is like living a battlefield. \\
    \bottomrule
\end{tabular}
\caption{\label{corpus_examples}
Examples of annotated metaphors in CMDAG, separated by source types.
}
\end{table*}

\subsection{Data Collection}

In constructing our corpus, we first collect a raw set of $\sim$153K probable metaphoric sentences from various Chinese literary sources online, with a focus on genres such as prose \footnote{https://www.ppzuowen.com/book/sanwen/}, poems \footnote{https://github.com/yuxqiu/modern-poetry}, and song \footnote{https://github.com/dengxiuqi/ChineseLyrics} and rap/hip-hop \footnote{https://github.com/djwackey/chinese-hiphop-lyrics} lyrics, which are often renowned for their rich usage of literary techniques and devices. Statistics of our raw and annotated metaphor datasets, separated by source types, are shown in Tables \ref{corpus_sources} and \ref{corpus_stats}. We applied the following set of heuristic rules to detect sentences which are likely to be of metaphoric usage, as opposed to literal ones, if either:
\begin{itemize}
    \item The sentence contains Chinese simile comparators ("像", "好似", "如同", etc.), or
    
    \item We identify metaphors by applying a similar method as in \citet{su2017}, where the sentence is classified as metaphoric if its subject and object, identified through dependency parsing, are not highly related and do not have a hyponym/hypernym relationship. We query whether the subject is a hyponym or a hypernym of the object in WordNet, and determine the relatedness between the subject and object by computing their cosine similarity score (a low score indicates the subject and object are less related, and hence there exists little shared information between them). 
    
    Suppose the subject and object are represented by $n$-dimensional vectors $\textbf{w}$ and $\textbf{v}$ respectively, then their cosine similarity score is computed as:
    \begin{equation*}
        \cos(\textbf{w}, \textbf{v}) = \frac{\sum_{i=1}^n w_iv_i}{\sqrt{\sum_{i=1}^n w_i^2}\sqrt{\sum_{i=1}^n v_i^2}}
    \end{equation*}
    where sentences with a score below a set threshold of $\cos(\textbf{w}, \textbf{v}) \leq 0.575$ (which from the results by \citet{su2017} gives the best performance and accuracy) are considered likely-metaphors and are kept for annotation.
\end{itemize}

\begin{table}
\centering
\footnotesize 
\setlength\tabcolsep{5pt}
\begin{tabular}{lc}
    \toprule
    \textbf{Source Type} & \textbf{\thead{Average Context \\Length (Tokens)}} \\
    \midrule
    Prose/Poem & 101 \\
    Song Lyrics & 49 \\
    Contemporary Poem & 51 \\
    HipHop/Rap Lyrics & 52 \\
    \midrule
    \textbf{Overall} & 59 \\
    \bottomrule
\end{tabular}
\caption{\label{corpus_stats}
Statistics of CMDAG and its raw data collection literature sources.
}
\end{table}

\subsection{Data Annotation}
\begin{figure*}
    \centering
    \begin{tabular}{cccc}
        \includegraphics[scale=0.3]{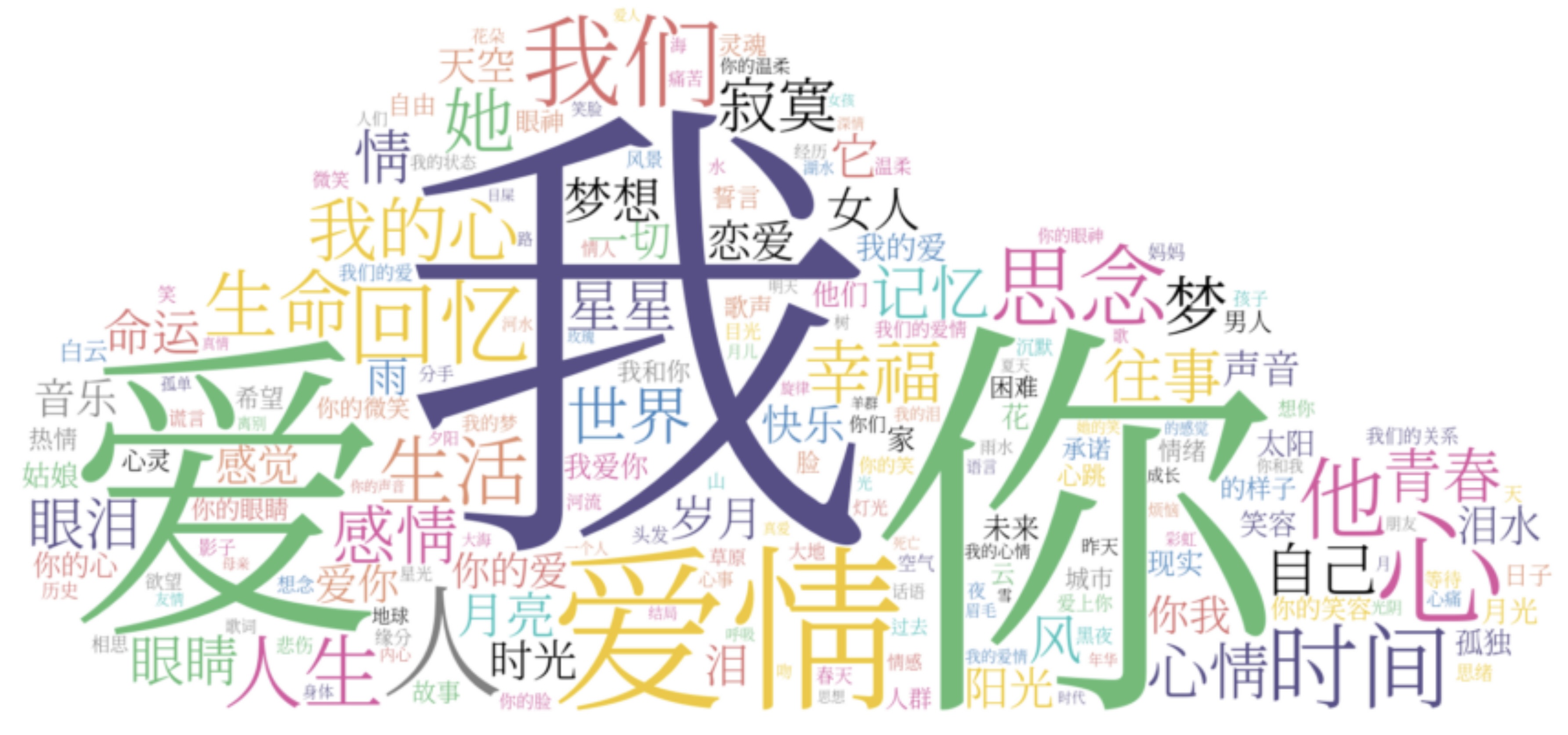} & 
        \includegraphics[scale=0.3]{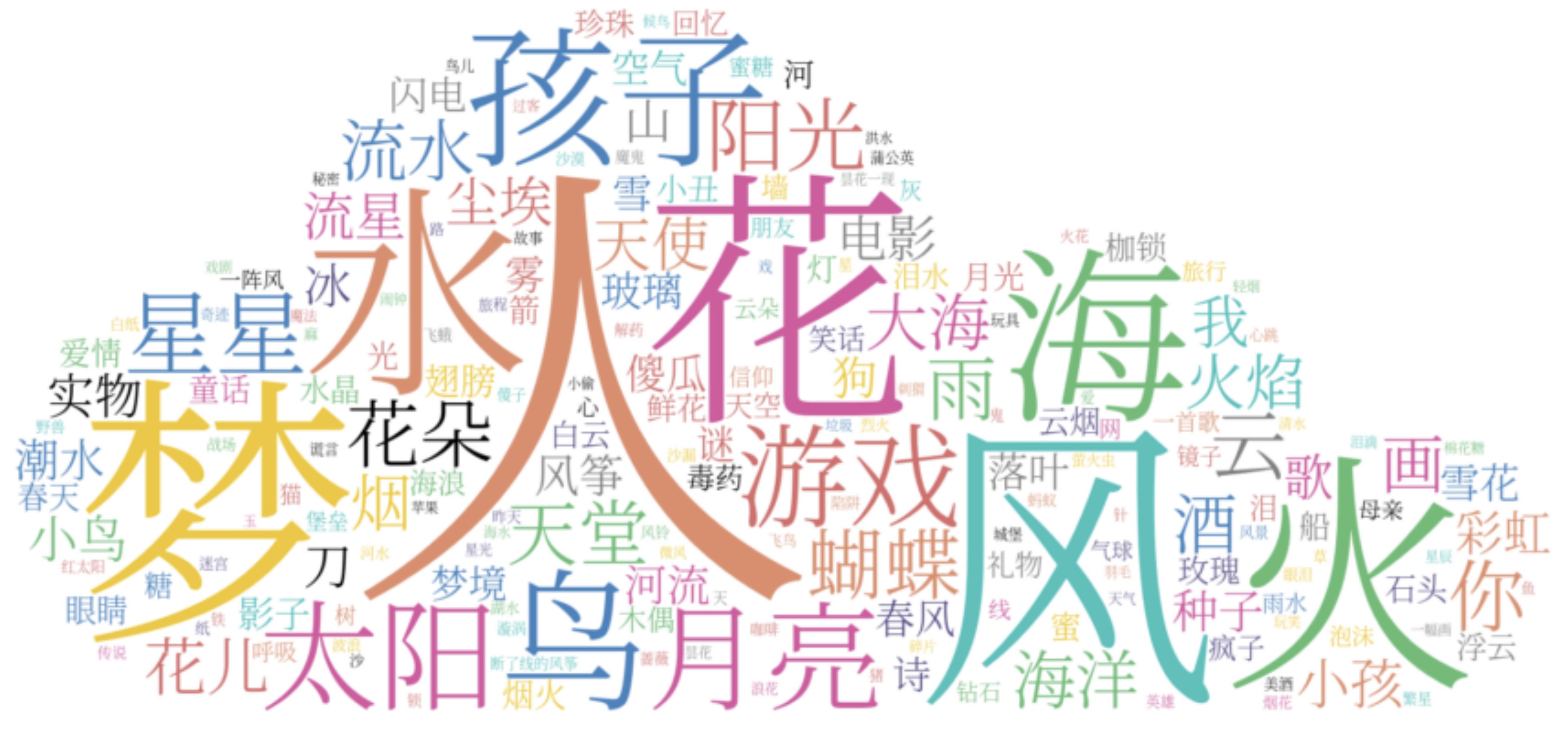} & 
        \includegraphics[scale=0.3]{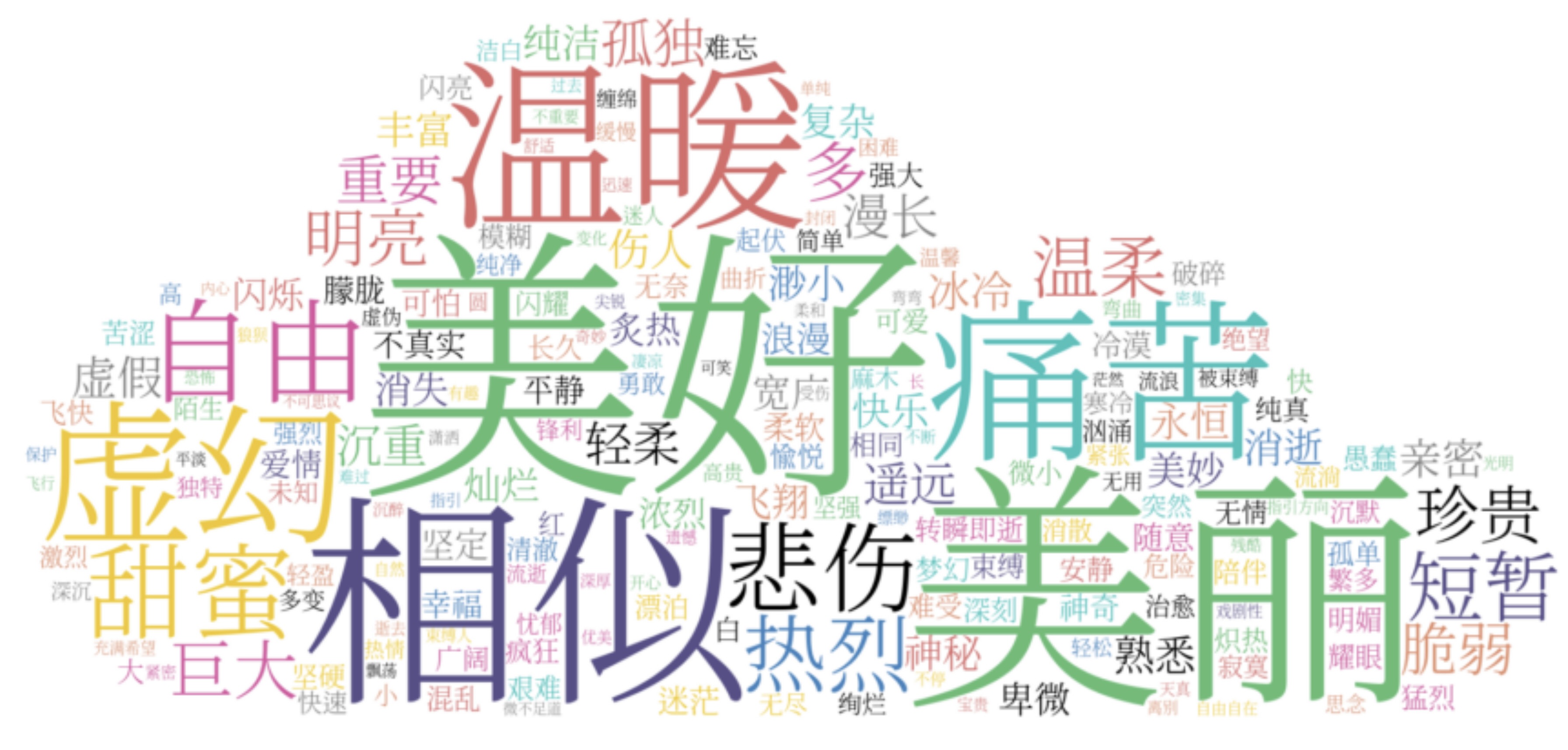} & 
        \includegraphics[scale=0.3]{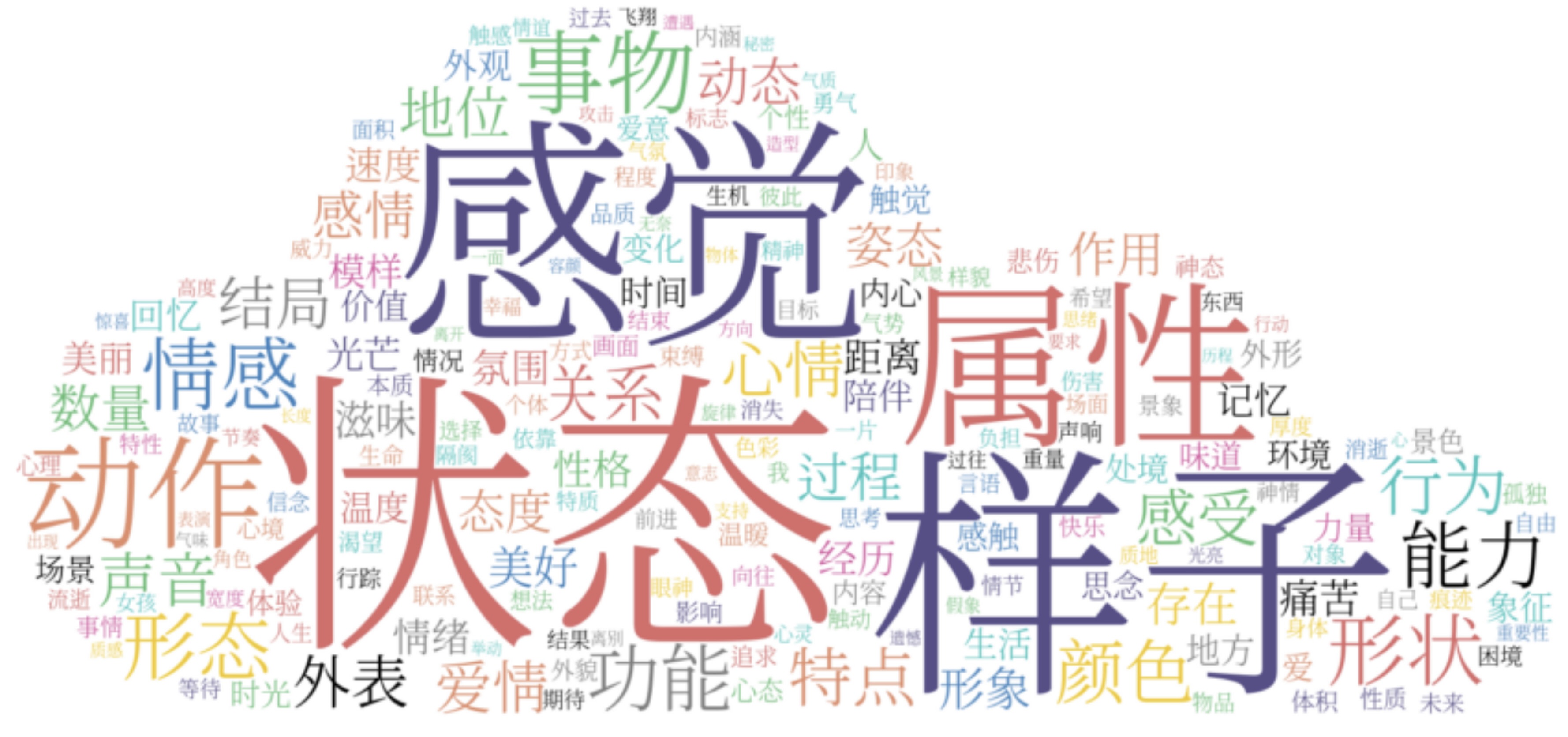} \\
        
        \includegraphics[scale=0.3]{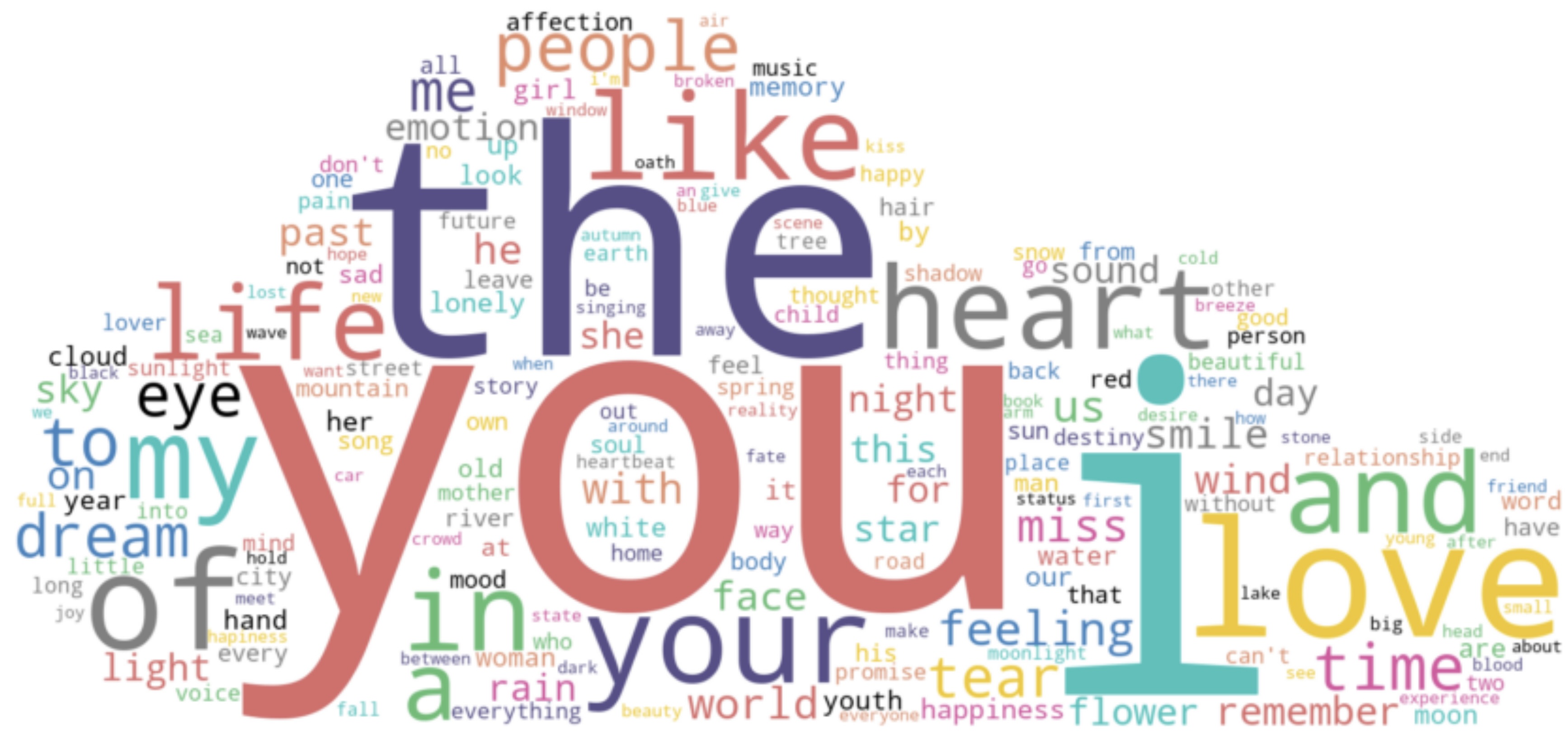} & 
        \includegraphics[scale=0.3]{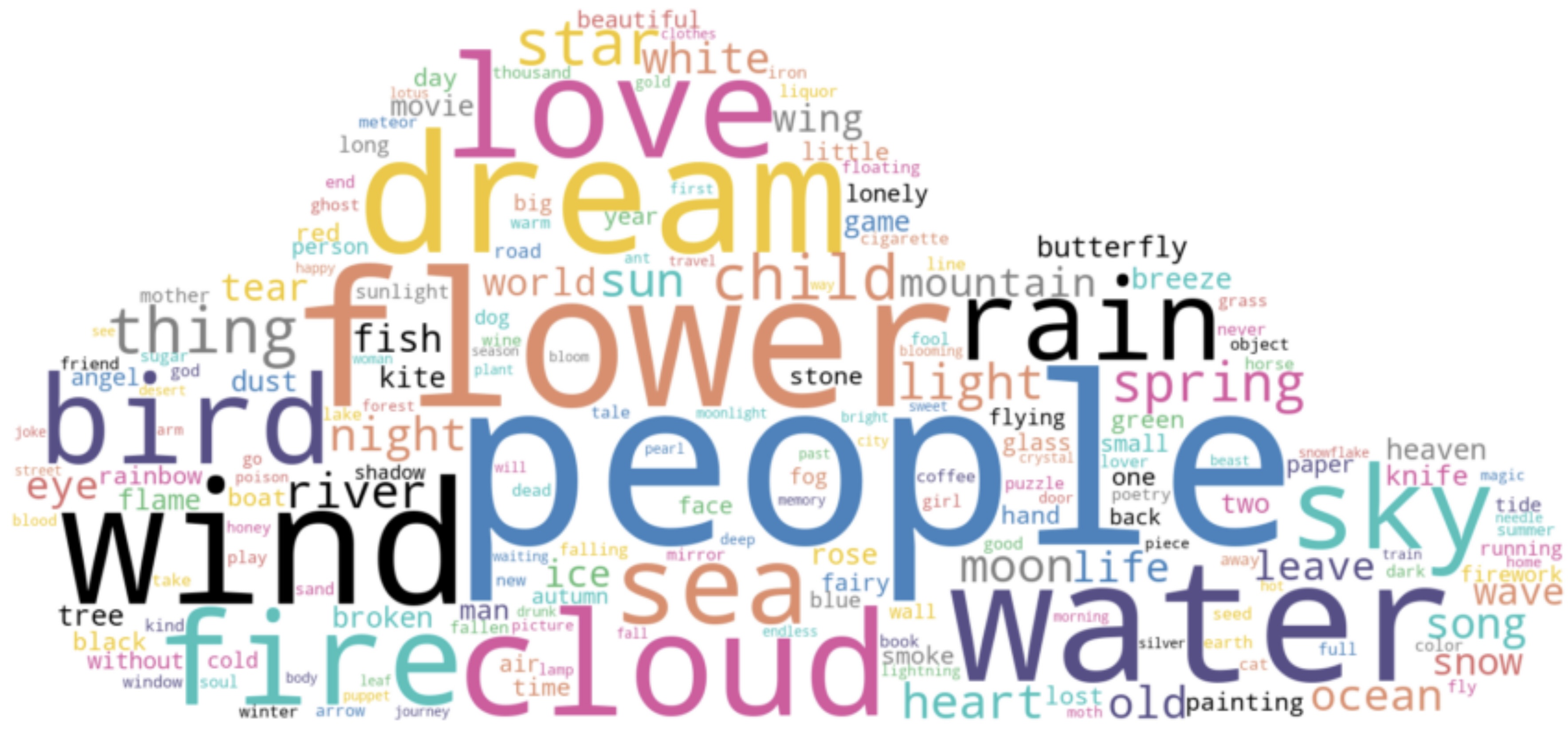} & 
        \includegraphics[scale=0.3]{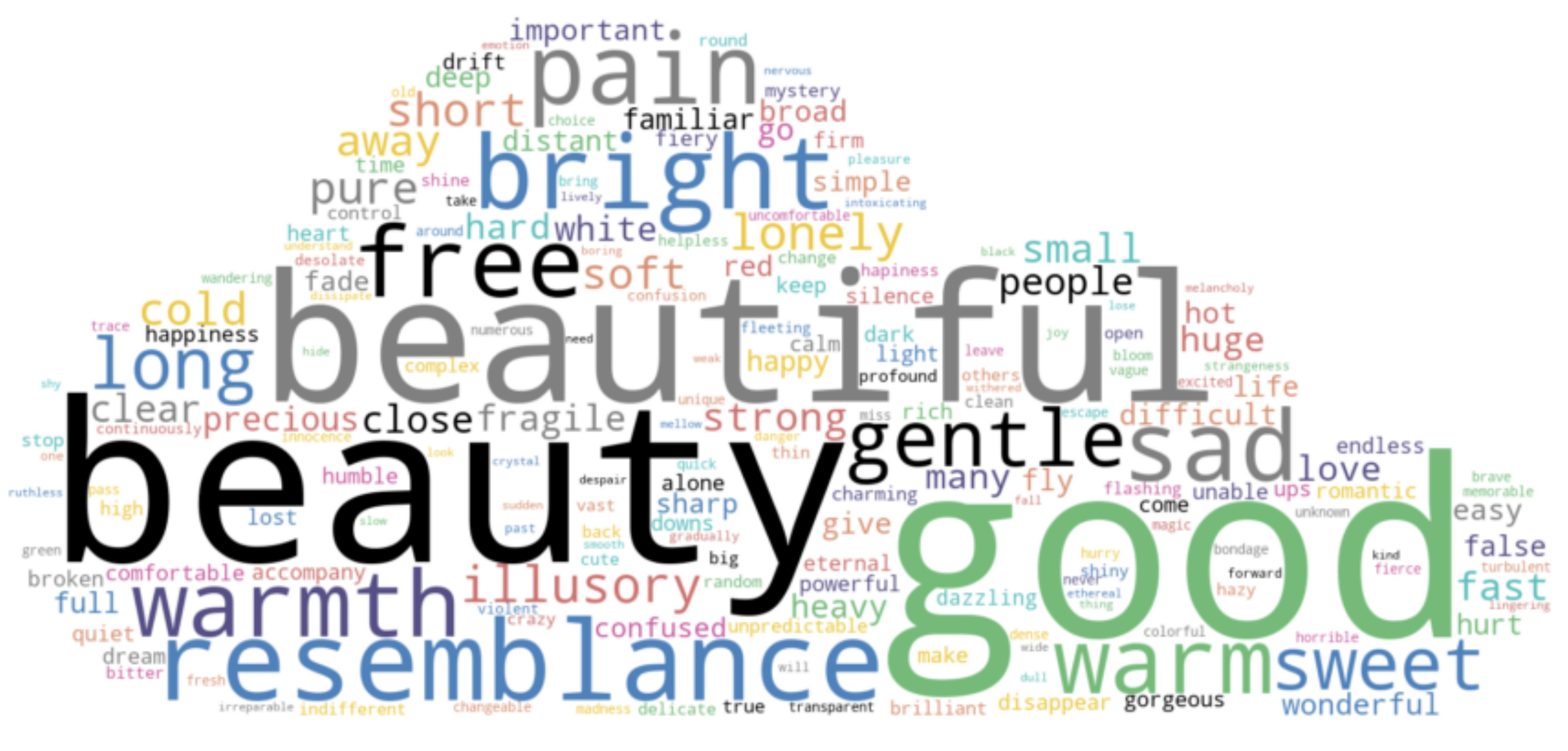} & 
        \includegraphics[scale=0.3]{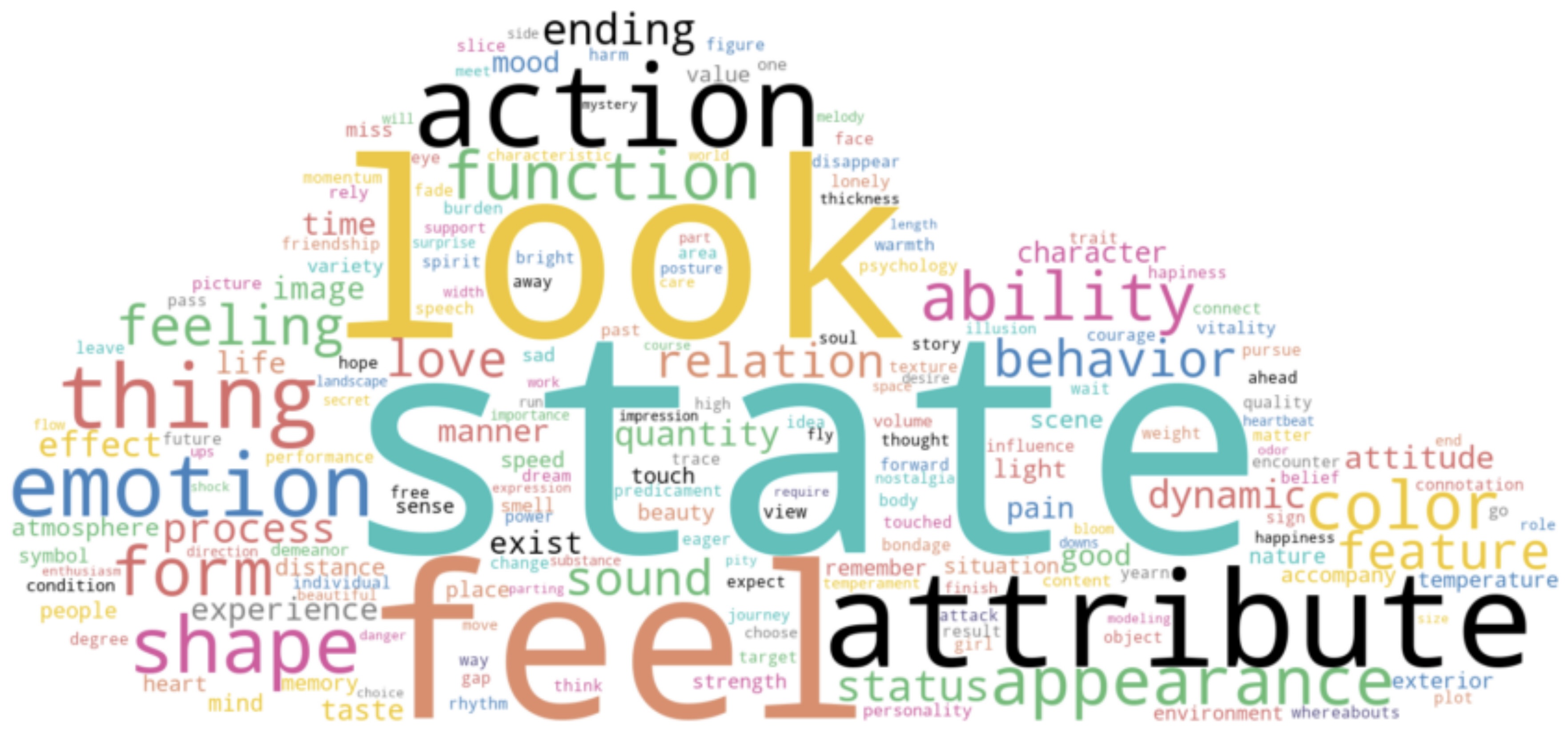} \\

        \textbf{Tenor} & \textbf{Vehicle} & \textbf{Ground (Adjective)} & \textbf{Ground (Noun)} \\
    \end{tabular}
    \caption{Word Clouds of tenors, vehicles, and adjective and noun components of grounds; the corresponding English word clouds are in the lower row.}
    \label{fig:word_cloud}
\end{figure*}

Our data annotation goal is to accurately mark each metaphor with a well-defined tuple of features: (\textsmaller{TENOR}, \textsmaller{VEHICLE}, \textsmaller{GROUND}). 

Consider the metaphor ``天上的云像奔腾的骏马'' (translates to: "clouds in the sky are like galloping horses"), annotated as: (云 (clouds), 奔腾的骏马 (galloping horses), 相似的形态 (similar forms)). A compilation of examples from our annotated dataset is presented in Table \ref{corpus_examples}.

Our annotation process consists of two main stages, focusing on both coarse-level and fine-level annotations:

\noindent \textbf{Preliminary Annotation}: Here, we engage a 20-people team of Chinese college students to identify genuine metaphors from the initial dataset, and highlight potential \textsmaller{TENOR}'s and \textsmaller{VEHICLE}'s.
Each sample is annotated by two annotators at this stage.

\noindent \textbf{Refined Annotation}: Leveraging the groundwork from the initial round, a second cohort of annotators, primarily Chinese native speakers with at least undergraduate credentials in Chinese Literature, refines the annotations.
Their specialized background enables them to further pinpoint the \textsmaller{GROUND} of the metaphors with higher precision.
We provide comprehensive guidelines to ensure consistent annotation quality, which mandates that each data piece is assessed by at least three annotators, improving label consistency and accuracy. Our labeling strategy emphasizes sophisticated composition of our \textsmaller{GROUND} labels, ensuring a structure combining an \textit{Adjective} and a \textit{Noun} (\textit{形容词} + \textit{名词}). 
The noun part delineates the shared characteristic between the \textsmaller{TENOR} and \textsmaller{VEHICLE}, while the adjective highlights the dimension underscoring their connection. 
Fig.~\ref{fig:word_cloud} showcases the diverse adjectives and noun elements of our annotated grounds via word clouds.

A cornerstone of our labeling strategy is the sophisticated composition of our \textsmaller{GROUND} labels. We ensure that they consistently adopt a structure melding both an \textit{Adjective} and a \textit{Noun} (\textit{形容词} + \textit{名词}). 
Specifically, the noun portion delineates the shared characteristic linking the \textsmaller{TENOR} and \textsmaller{VEHICLE}. Meanwhile, the accompanying adjective furnishes the dimension or aspect underscoring their connection. 
Fig.~\ref{fig:word_cloud} showcases the diverse adjectives and noun elements of our annotated grounds via word clouds. 

\noindent \textbf{Guidelines for Refined Annotation in Chinese}:
Our rigorous annotation approach is demonstrated through strict guidelines for the second annotation round, focusing on intricate annotation of metaphorical components in Chinese text.

1) Annotation and Quality Inspection Rules: 
Given Chinese rhetoric's complexity, it's essential to label all rhetorical devices like metaphor, metonymy, simile, personification, etc., in a unified standard. 
Annotators reference prior annotations, remaining cautious against possible inaccuracies, especially regarding previous GROUND labels, as we standardize the formatting requirements in the second round. 
A large proportion of statements contain multiple possible tuples of (TENOR, VEHICLE, GROUND). 
Annotators separate different tenors, vehicles, and grounds by three predetermined quotation marks when labeling each statement. 
Correctness verified, multiple tuples of one statement are automatically retrieved by string matching, forming the current open-sourced corpora.

2) Selection of Nouns for Grounds: 
To uphold annotation authenticity and precision, a curated list of nouns is provided. 
The emphasis is on opting for more descriptive nouns, avoiding generic terms like 样子 (appearance), 特征 (feature), 特点 (characteristic), 感受 (feeling), 感觉 (sensation), and other similar broad terms. 
This curated list is crucial for ensuring that the grounds aptly reflect the nuanced connections between the \textsmaller{TENOR} and \textsmaller{VEHICLE}.

\section{Methodology}

\begin{figure*}
    \centering
    \includegraphics[scale=0.15]{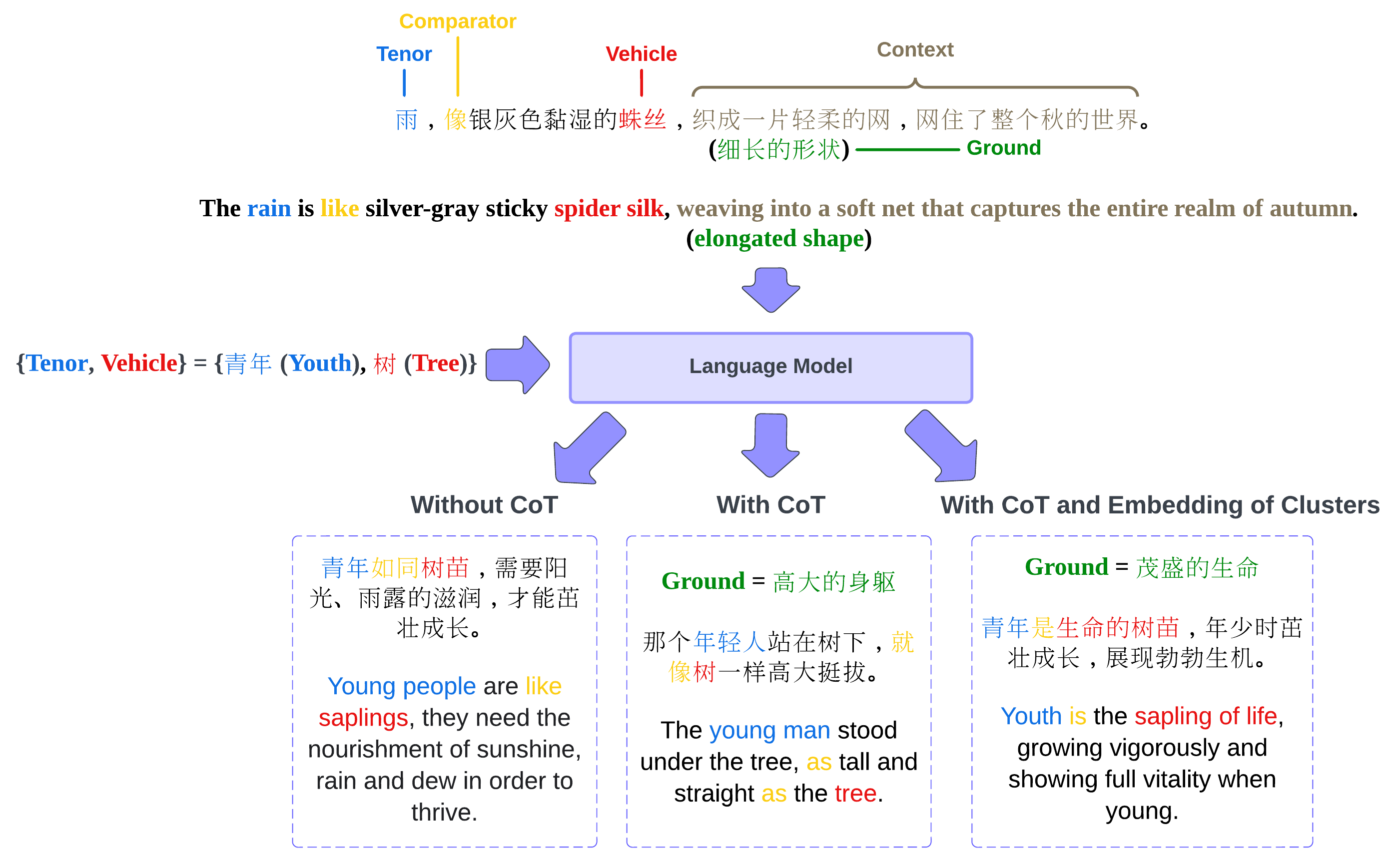}
    \caption{A flowchart that illustrates our experiment with an example of task 1.}
    \label{experiment_flowchart}
\end{figure*}

We consider two common scenarios that people often encounter with metaphor usage in writing. By utilizing our metaphor dataset with annotated grounds and applying a Chain-of-Thought (CoT) prompting technique with generated knowledge, we examine the importance of \textsmaller{GROUND} labels in metaphor generation tasks, including

\noindent \textbf{Task 1: Ground Identification}
The first task that we consider is a situation where given a potential pair of \textsmaller{TENOR} and \textsmaller{VEHICLE} as the subject and object which we would like to connect and compare, the model is to generate a corresponding metaphor.
\noindent \textbf{Task 2: Vehicle Identification}
Our second task requires the model to produce a metaphor when provided with a \textsmaller{TENOR} as the topic and a potential \textsmaller{GROUND} that signifies the features of the \textsmaller{TENOR} we aim to emphasize.

\subsection{Prompt Engineering}

With multiple-prompt prompting, we motivate in-context learning (ICL) by first of all providing the model with several examples (as in few-shot learning), utilizing the labels from our annotated dataset. For instance, based off our annotation of the metaphor ``他的温柔像海洋'' ("his gentleness is like the ocean"), an example prompted for task 1 would be: \textit{when constructing a metaphor with "his gentleness" as the tenor and "ocean" as the vehicle, the ground could be "the ability to accommodate" (``包容的能力'')}; and similarly an example can be prompted for task 2, with inferring the vehicle from the tenor and the ground.

\noindent \textbf{Chain-of-Thought (CoT) Prompting}
Using the model's response from the first prompt, we then ask the model to generate a metaphor from the \textsmaller{TENOR}-\textsmaller{VEHICLE} or \textsmaller{TENOR}-\textsmaller{GROUND} pairs, based on the \textsmaller{GROUND} or \textsmaller{VEHICLE} of its previous CoT generated response.

\noindent \textbf{Clustering}
To ensure a diverse set of examples, we strategically selected them from distinct clusters generated through various clustering algorithms. The initial clustering methodology leverages the embeddings of the \texttt{[CLS]} tokens to produce distinct clusters. 
Two primary clustering techniques were employed in our study:

\begin{enumerate}
    \item \textit{Sentence-level Embeddings Clustering:} This method utilizes the embeddings from the \texttt{[CLS]} token of each input, and the K-means clustering algorithm is then applied to these embeddings to generate distinct clusters.
    \item \textit{Word-level Embeddings Clustering:} Rather than using sentence-level embeddings, this technique takes advantage of word-level embeddings for each token in the input. These embeddings are then subjected to K-means clustering to produce the desired clusters.
\end{enumerate}
\section{Experiments}


As described in previous sections, we apply the unannotated and annotated versions of CMDAG. 

\subsection{Experimental Setting} 
\label{sec: experiment settings}
The evaluations were consistently carried out across six standardized settings to maintain a uniform benchmark, three for each of our two metaphor generation tasks, across our selected models. For each language model:
\begin{itemize}
    \item In Setting 0  of Task 1, we prompt the model with \textsmaller{TENOR}-\textsmaller{VEHICLE} pairs and for each pair we ask it to generate a corresponding metaphor.
    
    \item In Setting 1 of Task 1, we prompt the model with \textsmaller{TENOR}-\textsmaller{VEHICLE} pairs, as well as annotated examples selected based on our first clustering method, and for each pair we ask it to generate a corresponding \textsmaller{GROUND}. We then prompt the model again with the same \textsmaller{TENOR}-\textsmaller{VEHICLE} pairs and annotated examples, as well as the inferred \textsmaller{GROUND}, and for each pair we ask it to generate a corresponding metaphor.
    
    \item In Setting 2  of Task 1, we conduct a similar process as in Setting 1, except we select the annotated examples based on our second clustering method.

    \item In Task 2, we apply a similar procedure of settings as in Task 1, but instead of prompting with \textsmaller{TENOR}-\textsmaller{VEHICLE} pairs, we prompt the model and provide annotated examples with \textsmaller{TENOR}-\textsmaller{GROUND} pairs, and ask it to infer the corresponding \textsmaller{VEHICLE} for each pair in Settings 1 and 2.
\end{itemize}

\begin{table*}
\centering
\setlength\tabcolsep{5pt}
\begin{tabular}{lccccc}
\toprule
\textbf{Model Name} & \textbf{Setting} & \textbf{Clarity} & \textbf{Creativity} & \textbf{Authentic Expression} & \textbf{Final Score} \\
\midrule

Baichuan & $\bm{\odot}$ & 2.94 & 2.06 & 2.36 & 2.4 \\
Baichuan & $\bm{\diamondsuit}$ & 2.98 & 2.09 & 2.29 & 2.49 \\
Baichuan & $\bm{\star}$ & 2.98 & 2.07 & 2.20 & 2.32 \\

\midrule
Belle & $\bm{\odot}$ & 2.61 & 1.71 & 2.18 & 2.07 \\
Belle & $\bm{\diamondsuit}$ & 2.83 & 1.9 & 2.37 & 2.33 \\
Belle & $\bm{\star}$ & 2.97 & 1.69 & 2.23 & 2.17 \\

\midrule
GPT-4 & $\bm{\odot} $ & 2.92 & 1.64 & 2.16 & 2.25 \\
GPT-4 & $\bm{\diamondsuit} $ & 2.96 & 1.6 & 2.11 & 2.21 \\
GPT-4 & $\bm{\star} $ & 2.98 & 1.66 & 2.24 & 2.36 \\

\midrule
GPT-3.5 & $ \bm{\odot} $ & 2.99 & 1.78 & 2.23 & 2.21 \\
GPT-3.5 & $\bm{\diamondsuit} $ & 2.99 & 1.75 & 2.16 & 2.25 \\
GPT-3.5 & $ \bm{\star} $& 2.98 & 1.45 & 1.94 & 2.03 \\

\midrule
Chinese-alpaca-33B & $\bm{\odot} $ & 2.99 & 1.83 & 2.14 & 2.28 \\
Chinese-alpaca-33B & $\bm{\diamondsuit} $ & 2.97 & 1.68 & 2.14 & 2.11 \\
Chinese-alpaca-33B & $\bm{\star} $ & 2.99 & 1.86 & 2.29 & 2.20 \\

\midrule
ERNIE  & $\bm{\odot} $ & 2.87 & 1.86 & 2.30 & 2.27 \\
ERNIE  & $\bm{\diamondsuit} $ & 2.97 & 1.56 & 2.16 & 2.27 \\
ERNIE  & $\bm{\star} $ & 2.90 & 1.73 & 2.02 & 2.17 \\

\bottomrule

\end{tabular}
\caption{\label{tab:result task1} The human evaluation results for each model under three settings for taks1. According to the section~\ref{sec: experiment settings}, $\odot$ is the symbol of Setting 0, $\bm{\diamondsuit}$ is the symbol of Setting 1 and $\bm{\star}$ represents the Setting 2. }
\end{table*}

\begin{table*}
\centering

\setlength\tabcolsep{5pt}
\begin{tabular}{lccccc}
\toprule
\textbf{Model Name} & \textbf{Setting} & \textbf{Clarity} & \textbf{Creativity} & \textbf{Authentic Expression} & \textbf{Final Score} \\
\midrule

Baichuan & $ \bm{\odot} $ & 2.74 & 1.91 & 2.30 & 2.27 \\
Baichuan & $ \bm{\diamondsuit} $ & 2.41 & 1.89 & 1.98 & 2.04 \\
Baichuan & $ \bm{\star} $ & 2.53 & 1.98 & 2.00 & 2.08 \\

\midrule
Belle & $ \bm{\odot} $ & 2.47 & 1.93 & 2.29 & 2.14 \\
Belle & $ \bm{\diamondsuit} $ & 2.54 & 1.97 & 2.15 & 2.22 \\
Belle & $ \bm{\star} $ & 2.56 & 1.87 & 2.02 & 2.09 \\

\midrule
GPT-4 & $ \bm{\odot} $ & 2.57 & 1.83 & 2.17 & 2.13 \\
GPT-4 & $ \bm{\diamondsuit} $ & 2.48 & 1.66 & 2.26 & 2.12 \\
GPT-4 & $ \bm{\star} $ & 2.58 & 1.62 & 2.21 & 2.07 \\

\midrule
GPT-3.5 & $\bm{\odot} $ & 2.60 & 1.94 & 2.22 & 2.18 \\
GPT-3.5 & $\bm{\diamondsuit} $ & 2.36 & 1.77 & 2.12 & 2.02 \\
GPT-3.5 & $\bm{\star} $ & 2.49 & 1.70 & 2.05 & 2.02 \\

\midrule
ERNIE  & $\bm{\odot} $ & 2.23 & 1.85 & 2.21 & 1.98 \\
ERNIE  & $\bm{\diamondsuit}$ & 2.31 & 1.47 & 2.14 & 1.92 \\
ERNIE  & $\bm{\star} $ & 2.28 & 1.45 & 2.15 & 1.87 \\

\bottomrule

\end{tabular}
\caption{\label{tab:result task2} The human evaluation results for each model under three settings for taks2. The symbols for the settings are consistent with those in Table~\ref{tab:result task1}. }
\end{table*}

\begin{table}
\centering

\setlength\tabcolsep{5pt}
\begin{tabular}{lcrr}
\toprule
\textbf{Model Name} & \textbf{Setting} & \textbf{Task1} & \textbf{Task2} \\
\midrule
Belle & $\bm{\odot} $ & 0.112 & 0.236 \\
Belle & $\bm{\diamondsuit} $ & 0.12 & 0.268 \\
Belle & $\bm{\star} $ & 0.14 & 0.216 \\
\midrule
GPT-4 & $\bm{\odot} $ & 0.38 & 0.484 \\
GPT-4 & $\bm{\diamondsuit} $ & 0.448 & 0.548 \\
GPT-4 & $\bm{\star} $ & 0.448 & 0.548 \\
\midrule
GPT-3.5 & $\bm{\odot} $ & 0.372 & 0.384  \\
GPT-3.5 & $\bm{\diamondsuit} $ & 0.392 & 0.416  \\
GPT-3.5 & $\bm{\star} $ & 0.32 & 0.368  \\
\midrule

\end{tabular}
\caption{\label{tab:percentage of metaphors} Percentage of model-generated sentences that are reasonable Chinese metaphors. The symbols for the settings are consistent with those in Table~\ref{tab:result task1}.}
\end{table}
\subsubsection{Models}
Chinese metaphor generation is a novel task, we select three general generative models,  a Chinese nominal metaphor generation method, and a Chinese metaphor generation model as baselines.

\noindent \textbf{GPT-3.5 } GPT-3.5 \cite{openai2023} is a version of OpenAI's Generative Pretrained Transformer series. It is capable of handling a variety of language-processing tasks.

\noindent \textbf{GPT-4.0 } GPT-4 \cite{OpenAI2023GPT4TR} is a large-scale, multimodal model capable of accepting both image and text inputs to produce text outputs. It showcases human-level performance on various professional and academic benchmarks.

\noindent \textbf{Belle } Belle \cite{belle2023exploring} is a Chinese LLM (Large Language Model) trained specifically on Chinese data and thus is able to generate precise Chinese metaphoric information.

\noindent \textbf{Baichuan } Baichuan \cite{baichuan2023baichuan2} is a robust 13-billion parameter Chinese AI language model that is open-source and freely available for business and research purposes.

\noindent \textbf{Chinese-alpaca-33B }: Chinese-alpaca-33B \cite{chinese-llama-alpaca} is a state-of-the-art language model that holds a massive 33 billion parameters, specifically designed for Chinese language tasks.

\noindent \textbf{ERNIE }: Baidu ERNIE  \cite{baidu2023wenxin} is an innovative language model developed by Baidu Research, focusing on understanding and generating text in a more human-like manner. 

\subsubsection{Evaluation Metrics}
Evaluating models' performance on metaphor sentences is extremely challenging because determining the vividness of a metaphor is often intuitive. Many of these tasks cannot be measured by automatic metrics or even be judged by normal crowd workers. To get a more faithful evaluation, we hire expert annotators to judge model predictions.
All the annotators conducting the human evaluation have a Master's or Doctor's degree in Chinese Literature, Philology, or Literature. Due to cost, each sample is only analyzed by one annotator. To illustrate the annotators' responsibility, they are allowed to join the project only if their trial annotation results are verified by the authors of CMDAG.

The annotators are asked to rate the output based on whether it accurately and vividly generates the metaphors.
We implemented a four aspects rating system for categorizing the quality of the models’ outputs: \textbf{Clarity}, \textbf{Creativity}, \textbf{Authentic Expression} and \textbf{Final Score}. For every criterion, scores range from 1 to 3 points, with 1 being the minimum and 3 being the maximum score.

\noindent\textbf{Clarity:} Refers to the degree to which a statement is expressed without ambiguity, ensuring its comprehensibility.

\textit{Example:} 「眼睛是人心灵的窗户」This phrase, meaning "The eyes are the window to the soul", is unambiguous and clearly expresses the idea that one's eyes can reveal their innermost thoughts and feelings.

\noindent\textbf{Creativity:} Indicates the originality of the given statement, differentiating between novel concepts and clichéd ideas.

\textit{Example:} 「小朋友的脸仿佛是红苹果」This statement, which translates to "The faces of children are like red apples", is straightforward and lacks novelty.

\noindent\textbf{Authentic Expression:} Represents the degree to which a statement aligns with expressions that are considered authentic or native-like by the evaluators.

\textit{Example:} 「心如止水」This idiom, meaning "Heart like still water", is an authentic and native-like expression conveying a sense of inner peace and tranquility.

\noindent\subsection{Discussion}
We propose an analysis of how grounds-based CoT assists LLMs in metaphor generation in Tab.~\ref{tab:percentage of metaphors}.
Additionally, we provide expert-level human evaluation results on how different LLMs perform on Task 1 and Task 2 in Tab.~\ref{tab:result task1} and Tab.~\ref{tab:result task2}.
As supplementary material, we also reveal different human evaluation criteria' relationships in Tb.~\ref{tab:Pearson Relationship}.
The experiments of Tab.~\ref{tab:percentage of metaphors} are conducted on a selected 250-sample test set selected from CMDAG.
Only and all reasonable metaphorical sentences of various models and settings are manually evaluated and analyzed in Tab.~\ref{tab:result task1} and Tab.~\ref{tab:result task2}.

\subsubsection{Grounds-based CoT's Influence}
Tab.~\ref{tab:percentage of metaphors} reveals that Grounds-based CoT can improve the percentage of model-generated sentences that are reasonable Chinese metaphors.
Given Tab.~\ref{tab:result task1} and Tab.~\ref{tab:result task2}, we notice that LLMs with Grounds-based CoT achieve comparable performance on Task 1 and Task 2, compared with LLMs without Grounds-based CoT.
\textbf{Nota bene} that LLMs without grounds-based CoT often generate fewer reasonable metaphorical sentences, so their experiment results might slightly benefit from it.
We also propose that Grounds-based CoT leads to a slight performance decline, especially in the Creativity and Authentic Expression criteria.
An assumption of the observation is that Grounds-based CoT limits LLMs' tendency to explore novel \textbf{Vehicle (喻体)} and \textbf{Ground (喻意)}, which is a promising future research direction.

\subsubsection{Various LLMs' Performance}
Based on Tab.~\ref{tab:percentage of metaphors}, Tab.~\ref{tab:result task1}, and Tab.~\ref{tab:result task2}, we have two major observations. 
\textbf{First}, since Baichuan performs similarly or even surpasses GPT-4 and GPT-3.5 in Task 1 and Task 2, we point out that LLMs with more Chinese corpora in their pretraining procedure might perform better on Chinese metaphor generation.
\textbf{Second}, Belle generates much fewer reasonable metaphorical sentences compared to GPT-4 and GPT-3.5.
Additionally, GPT-4 and GPT-3.5 cannot always generate reasonable Chinese metaphorical sentences as well.
The observations reveal that the Chinese metaphor generation is still an under-explored task, and a larger model size and training corpus can lead to a noticeable performance gain on the task.

\subsubsection{Criteria's Relationships}
\begin{table}[h]
\footnotesize
\centering
\setlength\tabcolsep{5pt}
\begin{tabular}{lrrr}
\toprule
                  & \textbf{Clarity} & \textbf{Creativity} & \textbf{Authentic Expression} \\
\midrule
\textbf{Task 1}  & 0.28 & 0.71 & 0.68 \\
\textbf{Task 2}  & 0.85 & 0.72 & 0.41 \\
\bottomrule
\end{tabular}
\caption{ \label{tab:Pearson Relationship}Pearson Correlation between final score and evaluation criteria.}
\end{table}
Based on Tab.~\ref{tab:Pearson Relationship}, we point out that expert-level annotators attach importance to creativity when conducting an evaluation on Chinese metaphor generation.
Additionally, compared to the conventional Task 1, expert-level annotators pay more attention to clarity instead of authentic expression.
We propose an assumption that human annotators hold an implicit belief of 
\textsmaller{GROUND} in the conventional Task 1 setting which decreases their reliability on clarity.
However, writers practically only have \textsmaller{TENOR} and the features of \textsmaller{TENOR}, in other words \textsmaller{GROUND}, in their mind, when they want to write a metaphorical expression.
As a result, we point out that future metaphor generation models and benchmarks should pay more attention to the clarity of generated metaphorical sentences.

\section{Conclusion}

In this paper, we present an annotated Chinese Metaphor Dataset, encompassing approximately 28,000 sentences sourced from a wide array of Chinese literary forms, including poems, prose, and song lyrics. To ensure the precision and uniformity of our annotations, we have developed a thorough set of guidelines. These guidelines are instrumental in aiding annotators in the identification of tenors, vehicles, and grounds. Further more, we design a evaluation method for metaphor sentence generation that leverages a Chain of Thoughts (CoT) framework. Our experimental setup employs open-source multilingual Large Language Models (LLMs), which are tested to underscore the corpus's capability to facilitate the generation of creative and linguistically metaphors. This underscores the significant potential of our dataset to fuel advancements in the understanding and creation of Chinese metaphors.

\bibliography{acl_latex}

\begin{thebibliography}{31}
\expandafter\ifx\csname natexlab\endcsname\relax\def\natexlab#1{#1}\fi

\bibitem[{Baichuan(2023)}]{baichuan2023baichuan2}
Baichuan. 2023.
\newblock \href {https://arxiv.org/abs/2309.10305} {Baichuan 2: Open large-scale language models}.
\newblock \emph{arXiv preprint arXiv:2309.10305}.

\bibitem[{Black et~al.(1979)}]{black1979more}
Max Black et~al. 1979.
\newblock More about metaphor.
\newblock \emph{Metaphor and thought}, 2:19--41.

\bibitem[{Chakrabarty et~al.(2020)Chakrabarty, Muresan, and Peng}]{chakrabarty2020generating}
Tuhin Chakrabarty, Smaranda Muresan, and Nanyun Peng. 2020.
\newblock \href {http://arxiv.org/abs/2009.08942} {Generating similes effortlessly like a pro: A style transfer approach for simile generation}.

\bibitem[{Chakrabarty et~al.(2021)Chakrabarty, Zhang, Muresan, and Peng}]{chakrabarty2021mermaid}
Tuhin Chakrabarty, Xurui Zhang, Smaranda Muresan, and Nanyun Peng. 2021.
\newblock \href {http://arxiv.org/abs/2103.06779} {Mermaid: Metaphor generation with symbolism and discriminative decoding}.

\bibitem[{Chan et~al.(2023)Chan, Chen, Su, Yu, Xue, Zhang, Fu, and Liu}]{chan2023chateval}
Chi-Min Chan, Weize Chen, Yusheng Su, Jianxuan Yu, Wei Xue, Shanghang Zhang, Jie Fu, and Zhiyuan Liu. 2023.
\newblock Chateval: Towards better llm-based evaluators through multi-agent debate.
\newblock \emph{arXiv preprint arXiv:2308.07201}.

\bibitem[{Cui et~al.(2023)Cui, Yang, and Yao}]{chinese-llama-alpaca}
Yiming Cui, Ziqing Yang, and Xin Yao. 2023.
\newblock \href {https://arxiv.org/abs/2304.08177} {Efficient and effective text encoding for chinese llama and alpaca}.
\newblock \emph{arXiv preprint arXiv:2304.08177}.

\bibitem[{End(1986)}]{end1986grounds}
Laure~J End. 1986.
\newblock Grounds for metaphor comprehension.
\newblock In \emph{Advances in psychology}, volume~39, pages 327--345. Elsevier.

\bibitem[{Gong(2003)}]{gong2003corpus}
Shu-Ping Gong. 2003.
\newblock A corpus-based study on mapping principles of metaphors in politics.
\newblock In \emph{Proceedings of the ROCLING 2003 Student Workshop}, pages 287--294.

\bibitem[{Jiang et~al.(2023)Jiang, Li, Zhang, Huang, Lin, and Chen}]{jiang2023tigerscore}
Dongfu Jiang, Yishan Li, Ge~Zhang, Wenhao Huang, Bill~Yuchen Lin, and Wenhu Chen. 2023.
\newblock Tigerscore: Towards building explainable metric for all text generation tasks.
\newblock \emph{arXiv preprint arXiv:2310.00752}.

\bibitem[{Lakoff(1992)}]{lakoff1992}
George Lakoff. 1992.
\newblock The contemporary theory of metaphor.
\newblock In Andrew Ortony, editor, \emph{Metaphor and Thought (2nd edition)}, chapter~11, pages 202--251. Cambridge University Press, Cambridge.

\bibitem[{Li et~al.(2022{\natexlab{a}})Li, Lin, and Geurin}]{li2022nominal}
Yucheng Li, Chenghua Lin, and Frank Geurin. 2022{\natexlab{a}}.
\newblock \href {http://arxiv.org/abs/2206.05195} {Nominal metaphor generation with multitask learning}.

\bibitem[{Li et~al.(2022{\natexlab{b}})Li, Lin, and Guerin}]{li-etal-2022-nominal}
Yucheng Li, Chenghua Lin, and Frank Guerin. 2022{\natexlab{b}}.
\newblock \href {https://aclanthology.org/2022.inlg-main.18} {Nominal metaphor generation with multitask learning}.
\newblock In \emph{Proceedings of the 15th International Conference on Natural Language Generation}, pages 225--235, Waterville, Maine, USA and virtual meeting. Association for Computational Linguistics.

\bibitem[{Li et~al.(2023)Li, Wang, Lin, and Frank}]{li2023metaphor}
Yucheng Li, Shun Wang, Chenghua Lin, and Guerin Frank. 2023.
\newblock Metaphor detection via explicit basic meanings modelling.
\newblock \emph{arXiv preprint arXiv:2305.17268}.

\bibitem[{Lin(2021)}]{lin2021metaphor}
Su~Lin. 2021.
\newblock Metaphor and metonymy: Differences in chinese language and culture.
\newblock \emph{Open Journal of Modern Linguistics}, 11(2):135--139.

\bibitem[{Liu et~al.(2022)Liu, Jia, Zhang, Zhuang, Liu, and Vosoughi}]{liu2022second}
Ruibo Liu, Chenyan Jia, Ge~Zhang, Ziyu Zhuang, Tony Liu, and Soroush Vosoughi. 2022.
\newblock Second thoughts are best: Learning to re-align with human values from text edits.
\newblock \emph{Advances in Neural Information Processing Systems}, 35:181--196.

\bibitem[{Liu et~al.(2023)Liu, Yang, Jia, Zhang, Zhou, Dai, Yang, and Vosoughi}]{liu2023training}
Ruibo Liu, Ruixin Yang, Chenyan Jia, Ge~Zhang, Denny Zhou, Andrew~M Dai, Diyi Yang, and Soroush Vosoughi. 2023.
\newblock Training socially aligned language models in simulated human society.
\newblock \emph{arXiv preprint arXiv:2305.16960}.

\bibitem[{Liu et~al.(2019)Liu, Fu, Cao, de~Melo, Tam, Niu, and Zhou}]{liu-etal-2019-rhetorically}
Zhiqiang Liu, Zuohui Fu, Jie Cao, Gerard de~Melo, Yik-Cheung Tam, Cheng Niu, and Jie Zhou. 2019.
\newblock \href {https://doi.org/10.18653/v1/P19-1192} {Rhetorically controlled encoder-decoder for {M}odern {C}hinese poetry generation}.
\newblock In \emph{Proceedings of the 57th Annual Meeting of the Association for Computational Linguistics}, pages 1992--2001, Florence, Italy. Association for Computational Linguistics.

\bibitem[{OpenAI(2023{\natexlab{a}})}]{OpenAI2023GPT4TR}
OpenAI. 2023{\natexlab{a}}.
\newblock Gpt-4 technical report.
\newblock \emph{ArXiv}, abs/2303.08774.

\bibitem[{OpenAI(2023{\natexlab{b}})}]{openai2023}
OpenAI. 2023{\natexlab{b}}.
\newblock \href {https://help.openai.com/en/articles/6643408-how-do-davinci-and-text-davinci-003-differ} {How do davinci and text davinci-003 differ?}

\bibitem[{Research(2023)}]{baidu2023wenxin}
Baidu Research. 2023.
\newblock \href {http://research.baidu.com/Blog/index-view?id=165} {Wenxin: Baidu's advanced language model}.

\bibitem[{Stowe et~al.(2021)Stowe, Chakrabarty, Peng, Muresan, and Gurevych}]{stowe-etal-2021-metaphor}
Kevin Stowe, Tuhin Chakrabarty, Nanyun Peng, Smaranda Muresan, and Iryna Gurevych. 2021.
\newblock \href {https://doi.org/10.18653/v1/2021.acl-long.524} {Metaphor generation with conceptual mappings}.
\newblock In \emph{Proceedings of the 59th Annual Meeting of the Association for Computational Linguistics and the 11th International Joint Conference on Natural Language Processing (Volume 1: Long Papers)}, pages 6724--6736, Online. Association for Computational Linguistics.

\bibitem[{Su et~al.(2017)Su, Huang, and Chen}]{su2017}
Chang Su, Shuman Huang, and Yijiang Chen. 2017.
\newblock Automatic detection and interpretation of nominal metaphor based on the theory of meaning.
\newblock \emph{Neurocomputing}, 219:300--311.

\bibitem[{Wachowiak and Gromann(2023)}]{wachowiak2023does}
Lennart Wachowiak and Dagmar Gromann. 2023.
\newblock Does gpt-3 grasp metaphors? identifying metaphor mappings with generative language models.
\newblock In \emph{Proceedings of the 61st Annual Meeting of the Association for Computational Linguistics (Volume 1: Long Papers)}, pages 1018--1032.

\bibitem[{Wang et~al.(2023)Wang, Zhang, Yang, Shi, Zhou, Hao, Xiong, Li, Sim, Chen et~al.}]{wang2023interactive}
Zekun Wang, Ge~Zhang, Kexin Yang, Ning Shi, Wangchunshu Zhou, Shaochun Hao, Guangzheng Xiong, Yizhi Li, Mong~Yuan Sim, Xiuying Chen, et~al. 2023.
\newblock Interactive natural language processing.
\newblock \emph{arXiv preprint arXiv:2305.13246}.

\bibitem[{Wei et~al.(2022)Wei, Wang, Schuurmans, Bosma, Xia, Chi, Le, Zhou et~al.}]{wei2022chain}
Jason Wei, Xuezhi Wang, Dale Schuurmans, Maarten Bosma, Fei Xia, Ed~Chi, Quoc~V Le, Denny Zhou, et~al. 2022.
\newblock Chain-of-thought prompting elicits reasoning in large language models.
\newblock \emph{Advances in Neural Information Processing Systems}, 35:24824--24837.

\bibitem[{Yang et~al.(2023)Yang, Liu, Lei, Yang, Wei, Liu, and Xie}]{yang-etal-2023-fantastic}
Kexin Yang, Dayiheng Liu, Wenqiang Lei, Baosong Yang, Xiangpeng Wei, Zhengyuan Liu, and Jun Xie. 2023.
\newblock \href {https://doi.org/10.18653/v1/2023.acl-long.28} {Fantastic expressions and where to find them: {C}hinese simile generation with multiple constraints}.
\newblock In \emph{Proceedings of the 61st Annual Meeting of the Association for Computational Linguistics (Volume 1: Long Papers)}, pages 468--486, Toronto, Canada. Association for Computational Linguistics.

\bibitem[{Yue et~al.(2023)Yue, Qu, Zhang, Fu, Huang, Sun, Su, and Chen}]{yue2023mammoth}
Xiang Yue, Xingwei Qu, Ge~Zhang, Yao Fu, Wenhao Huang, Huan Sun, Yu~Su, and Wenhu Chen. 2023.
\newblock Mammoth: Building math generalist models through hybrid instruction tuning.
\newblock \emph{arXiv preprint arXiv:2309.05653}.

\bibitem[{Yunjie et~al.(2023)Yunjie, Yong, Yan, Yiping, Qiang, Lei, Baochang, and Xiangang}]{belle2023exploring}
Ji~Yunjie, Deng Yong, Gong Yan, Peng Yiping, Niu Qiang, Zhang Lei, Ma~Baochang, and Li~Xiangang. 2023.
\newblock Exploring the impact of instruction data scaling on large language models: An empirical study on real-world use cases.
\newblock \emph{arXiv preprint arXiv:2303.14742}.

\bibitem[{Zhang et~al.(2023)Zhang, Shi, Liu, Yuan, Li, Dong, Shu, Li, Wang, Lin et~al.}]{zhang2023chinese}
Ge~Zhang, Yemin Shi, Ruibo Liu, Ruibin Yuan, Yizhi Li, Siwei Dong, Yu~Shu, Zhaoqun Li, Zekun Wang, Chenghua Lin, et~al. 2023.
\newblock Chinese open instruction generalist: A preliminary release.
\newblock \emph{arXiv preprint arXiv:2304.07987}.

\bibitem[{Zhang et~al.(2021)Zhang, Cui, Xia, Guo, Li, Wei, and Cui}]{zhang2021writing}
Jiayi Zhang, Zhi Cui, Xiaoqiang Xia, Yalong Guo, Yanran Li, Chen Wei, and Jianwei Cui. 2021.
\newblock Writing polishment with simile: Task, dataset and a neural approach.
\newblock In \emph{Proceedings of the AAAI Conference on Artificial Intelligence}, volume~35, pages 14383--14392.

\bibitem[{Zheng et~al.(2020)Zheng, Song, Hu, Fu, and Zhou}]{zheng2020love}
Danning Zheng, Ruihua Song, Tianran Hu, Hao Fu, and Jin Zhou. 2020.
\newblock \href {http://arxiv.org/abs/2001.00733} {"love is as complex as math": Metaphor generation system for social chatbot}.

\end{thebibliography}
\end{CJK*}
\end{document}